\documentclass[conference]{IEEEtran}
\IEEEoverridecommandlockouts
\usepackage{cite}
\usepackage{amsmath,amssymb,amsfonts}
\usepackage{algorithmic}
\usepackage{graphicx}
\usepackage{textcomp}
\usepackage{xcolor}
 \usepackage{booktabs}
 \usepackage{float}
 \usepackage{listings}
 \usepackage{url}
 \usepackage{blindtext}
 \graphicspath{{./figures/}}
\def\BibTeX{{\rm B\kern-.05em{\sc i\kern-.025em b}\kern-.08em
    T\kern-.1667em\lower.7ex\hbox{E}\kern-.125emX}}
\begin{document}

\title{AI Agent for Reverse-Engineering Legacy Finite-Difference Code and Translating to Devito}

\author{
\IEEEauthorblockN{Yinghan Hou}
\IEEEauthorblockA{
Department of Earth Science and Engineering \\
Imperial College London \\
London, United Kingdom \\
ghwzhyinghan@gmail.com
}

\and
\IEEEauthorblockN{Zongyou Yang}
\IEEEauthorblockA{
Department of Computer Science \\
University College London \\
London, United Kingdom \\
dryang0624@gmail.com
}
}

\maketitle

\begin{abstract}
To facilitate the transformation of legacy finite difference implementations into the Devito environment, this study develops an integrated AI agent framework. Retrieval-Augmented Generation (RAG) and open-source Large Language Models are combined through multi-stage iterative workflows in the system's hybrid LangGraph architecture.
The agent constructs an extensive Devito knowledge graph through document parsing, structure-aware segmentation, extraction of entity relationships, and Leiden-based community detection. GraphRAG optimisation enhances query performance across semantic communities that include seismic wave simulation, computational fluid dynamics, and performance tuning libraries. A reverse engineering component derives three-level query strategies for RAG retrieval through static analysis of Fortran source code. To deliver precise contextual information for language model guidance, the multi-stage retrieval pipeline performs parallel searching, concept expansion, community-scale retrieval, and semantic similarity analysis.
Code synthesis is governed by Pydantic-based constraints to guarantee structured outputs and reliability. A comprehensive validation framework integrates conventional static analysis with the G-Eval approach, covering execution correctness, structural soundness, mathematical consistency, and API compliance.
The overall agent workflow is implemented on the LangGraph framework and adopts concurrent processing to support quality-based iterative refinement and state-aware dynamic routing. The principal contribution lies in the incorporation of feedback mechanisms motivated by reinforcement learning, enabling a transition from static code translation toward dynamic and adaptive analytical behavior.
\end{abstract}


\section{Introduction}
\subsection{Background and problem description}
High-performance scientific computing depends strongly on large-scale legacy Fortran codebases that support essential applications, including weather prediction, climate simulation, nuclear safety analysis, and computational physics. Such codes embody many decades of accumulated engineering knowledge and correspond to huge investments. However, the transition toward contemporary hardware platforms such as GPUs and multi-core processors, together with rapidly evolving software ecosystems, has made the effective reuse and integration of legacy implementations progressively more challenging. Meanwhile, the population of experienced Fortran developers is steadily shrinking, as the language has largely vanished from modern computer science education.
Current strategies for code modernisation exhibit notable shortcomings. Conventional tools such as F2PY are mainly oriented toward interface wrapping rather than deep restructuring. Fully manual refactoring can preserve accuracy, yet the associated time and financial costs are often unacceptable. Although large language models demonstrate potential for automated code translation, current general-purpose solutions lack a specialized focus on scientific computing and quality validation.
To overcome these issues, this project presents an AI-based agent system. The proposed system performs automated reverse engineering of legacy Fortran finite-difference programs and converts them into Devito, a contemporary domain-specific language designed for finite difference computations.

\subsection{GraphRAG}
Retrieval augmented generation (RAG) enhances the capabilities of large language models by incorporating external knowledge drawn from databases or document collections \cite{codellm2023}. GraphRAG, introduced by Microsoft Research, expands this paradigm through the integration of knowledge graph representations \cite{graphrag2024}. In this approach, graphs are constructed in which nodes correspond to entities or documents, while edges encode the relationships that connect them \cite{graphrag2024}. Information retrieval is carried out at the level of graph communities, enabling the combination of localized evidence with an overarching structural context. Such a mechanism is especially valuable when dealing with extensive and technically complex corpora. In the present study, GraphRAG is particularly appropriate because the Devito repository generates tens of thousands of structured components along with dense networks of entity relationships. The framework accelerates query execution through the application of community detection and graph sparsification strategies, both of which are essential for real-time retrieval scenarios. Moreover, GraphRAG also keeps code-level dependencies and semantic relations. These properties are well-matched to the task of reverse engineering Fortran into Devito, where precision and structural fidelity are critical \cite{raganything2025, croft2010, Robinson2015, langchain2023rag1}.

\begin{figure}[H]
\centering
\includegraphics[width=\linewidth,height=.75\textheight,keepaspectratio]{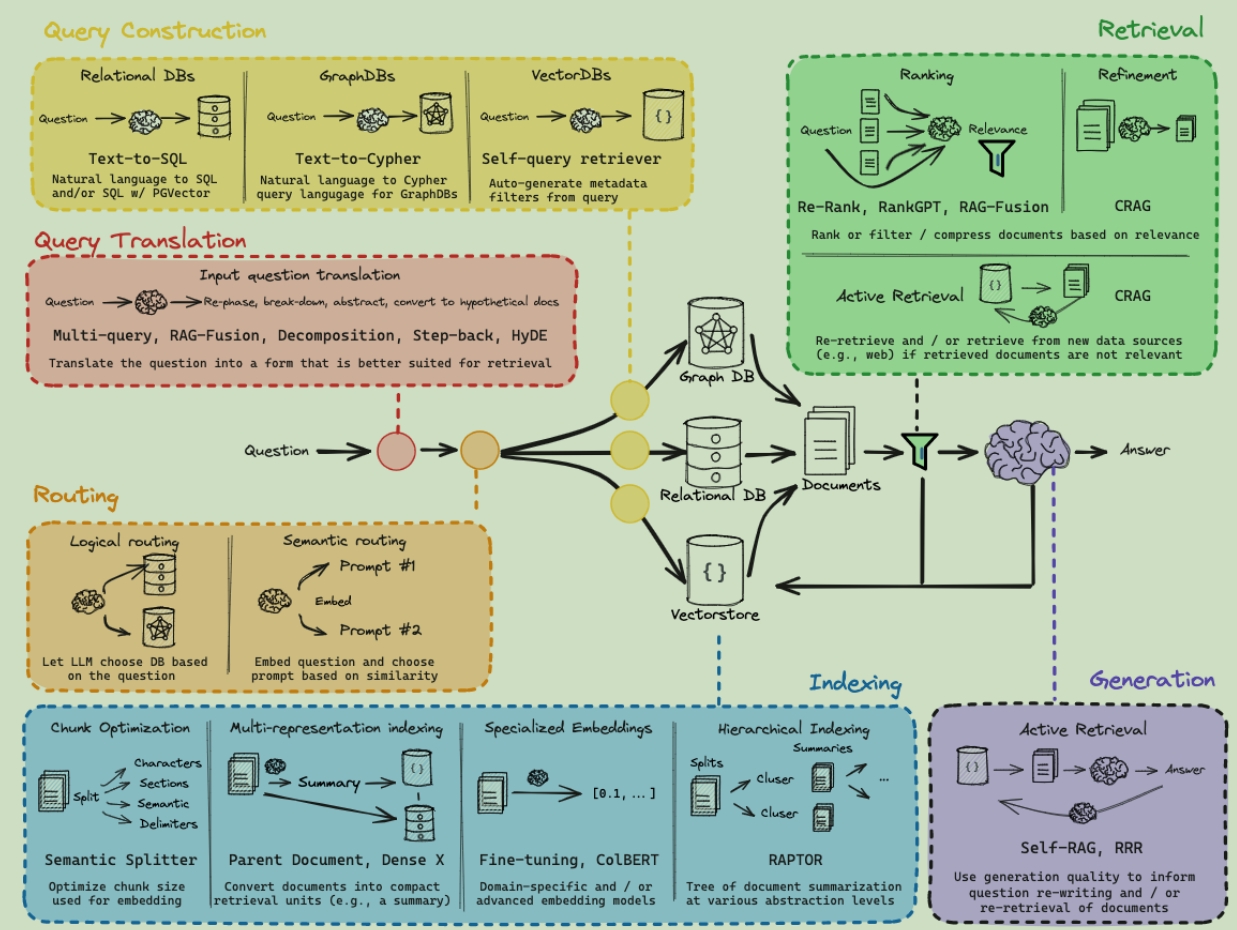}
\caption{LangChain RAG from scratch pipeline overview.}
\label{fig:rag_from_scratch}
\end{figure}

\subsection{Embeddings}

Embedding models transform textual content or source code into numerical vector representations that capture underlying semantic information. This transformation enables similarity computation and forms the foundation for effective retrieval mechanisms. In this study, the BGE-M3 embedding model is adopted \cite{chen2024bge}. This selection is motivated by its strong results on the MTEB benchmark, which assesses embedding approaches across a wide range of retrieval and clustering scenarios \cite{muennighoff2023mteb, reimers2019sentencebert}. BGE-M3 supports extended input lengths of up to 8,192 tokens and generates embeddings with 1024-dimensions. Such features are essential in this context, as technical documentation and code fragments are frequently lengthy and structurally complex.
The model is also efficient in terms of size and speed, making it suitable for deployment at large scales. BGE-M3 also provides multimodal capability, allowing both natural language text and programming code to be embedded within a unified vector space. This capability facilitates the construction of a cohesive knowledge base that unifies documentation, source code, and structured metadata. Consequently, BGE-M3 is a robust foundation for knowledge graph construction and supports dependable retrieval within the Fortran to Devito translation framework.

\subsection{K3-Trans}
\label{sec:k3trans}

K3-Trans is an approach to facilitate the development of structured knowledge bases for program translation tasks \cite{wang2024k3trans}. It focuses on three core components, including a library of target language examples, explicit mapping rules that link source and target languages, and a repository containing validated pairs of translated programs.
Within this project, the underlying concepts of K3-Trans are employed to bridge legacy Fortran finite difference codes with the Devito framework. The example collection supplies Devito implementations covering a range of numerical schemes and boundary condition settings. The mapping rules formalize the transformation from Fortran syntactic constructs to symbolic operators used in Devito. The set of translation pairs provides verified correspondences between scientific formulations and their Devito implementations. Through this approach, the translation process attains a high level of accuracy and reproducibility, which is essential for scientific computing applications that demand reliable and stable numerical performance.

\section{Methodology and Implementation}
\subsection{System architecture overview}
\begin{figure*}[t]
\centering
\includegraphics[width=\linewidth,height=.75\textheight,keepaspectratio]{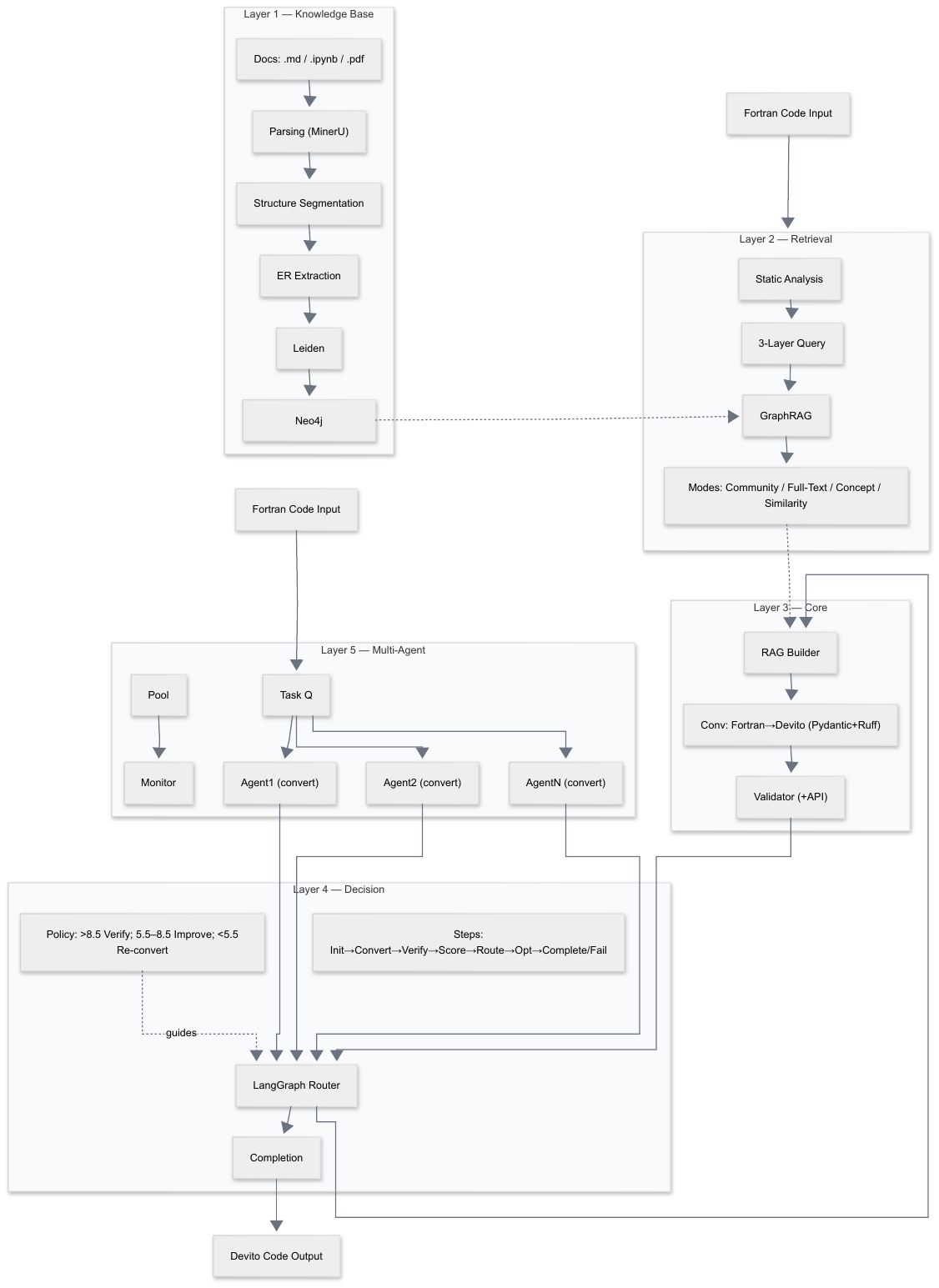}
\caption{System architecture for Fortran-to-Devito pipeline.}
\label{fig:fortran_devito_arch}
\end{figure*}

The system is organized around a five-layer hierarchical architectural design. Each layer communicates through standardized interfaces to support coordinated interaction across layers. The first layer is the Knowledge Base Construction Layer, which handles documents in multiple formats. Its functions include document parsing, structure-aware segmentation, entity relationship extraction, and Leiden-based community detection. The second layer is the Retrieval Enhancement Layer, where multimodal parallel retrieval is implemented. GraphRAG strategies are integrated with static analysis of Fortran code. Four parallel retrieval mechanisms provide accurate contextual information: community-level retrieval, full text search, concept expansion, and semantic similarity matching.
The third layer is the Core Components Layer, which contains the RAG Context Builder, the Fortran-Devito conversion module, and the Quality Validation component. Structured output is enforced through Pydantic-based constraints, while code quality checks are supported through Ruff integration. The fourth layer is the Decision Layer, which implements dynamic routing based on the LangGraph framework across eight workflow nodes. Quality-based adaptive decision-making guides the conversion process. The fifth layer is the Agent Coordination Layer, which supports parallel execution through intelligent task queue management. System robustness is ensured through unified error handling strategies, while computational efficiency is optimized through automatic scaling of agent resources.
The architectural flowchart shown in Figure \ref{fig:fortran_devito_arch} illustrates the complete processing pipeline from Fortran source input to Devito code generation, demonstrating data exchange among the five architectural layers, interactions between components, and the corresponding decision pathways.

\subsection{Knowledge base construction pipeline}
\subsubsection{Devito code repository analysis and preprocessing}

\begin{figure}[htbp]
\centering
\includegraphics[width=0.8\linewidth]{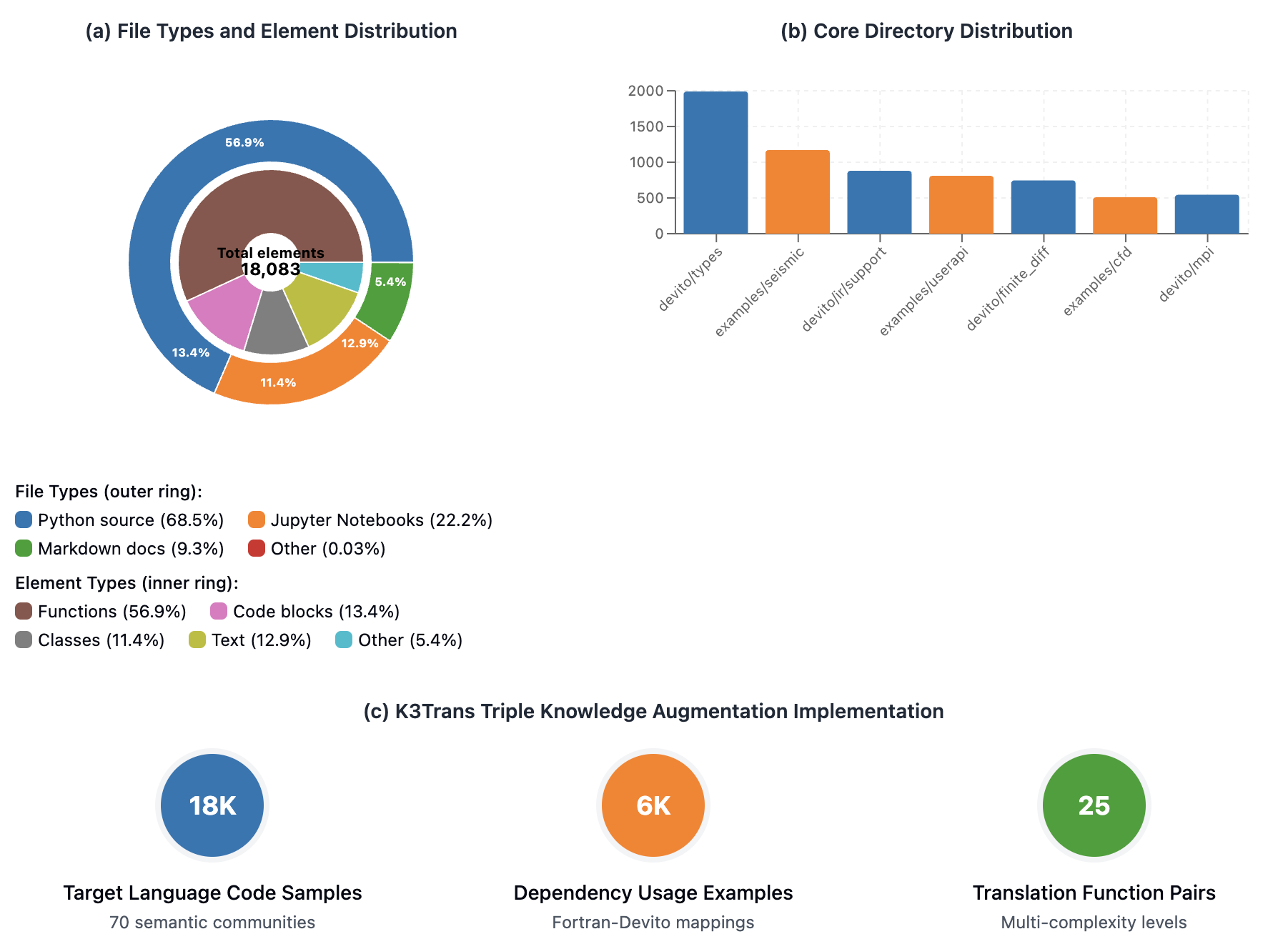}
\caption{Devito knowledge base}
\label{fig:devito_knowledge_base}
\end{figure}

The Devito repository analysis and preprocessing phase performs an extensive, structured examination of the original Devito codebase. This stage establishes a unified data foundation that supports subsequent document parsing and knowledge extraction procedures. Based on the K3Trans framework principles described in Section \ref{sec:k3trans}, the system integrates a triple knowledge enhancement strategy with the specific demands of scientific computing. Through this process, a dedicated knowledge base is developed to support Fortran-Devito translation tasks. Meanwhile, the project constructs a comprehensive library of validated translation pairs spanning multiple levels of complexity. Figure \ref{fig:devito_knowledge_base} presents the statistical characteristics of the Devito knowledge base derived according to the K3Trans framework principles.

\subsubsection{Multimodal document processing and MinerU integration}

The multimodal document processing stage provides structured parsing capabilities comparable to MinerU. It transforms originally unstructured documents into structured metadata streams. This stage supports Markdown files (\texttt{.md}), Jupyter Notebooks (\texttt{.ipynb}), PDF documents (\texttt{.pdf}), and Python source code. During parsing, textual content, source code segments, tables, and images are extracted while associated structural metadata is retained.
Distinct document formats are handled through dedicated processing strategies:
\begin{itemize}
\item \textbf{Python:} Abstract syntax tree (AST) parsing is employed to precisely identify program elements such as function definitions, class declarations, and documentation strings, thereby maintaining syntactic correctness.
\item \textbf{Jupyter Notebook:} Explicit attention is given to preserving the semantic relationship between executable code cells and documentation cells.
\item \textbf{Markdown:} Heading hierarchies are automatically detected, and complex structures including tables, lists, and embedded code blocks are systematically parsed.
\end{itemize}
Each resulting knowledge block is organized around a section heading and aggregates all associated text, tables, images, and code elements contained within that section. Through this process, 18,083 initially unstructured data items are transformed into structured components. Every component is annotated with metadata such as file location, content length, word count, and directory category.
The final structured metadata stream consists of 10,287 function elements, 2,423 code blocks, 2,065 class definitions, 2,339 text paragraphs, 600 headings, 249 lists, and 83 tables.

\subsubsection{Structure-aware semantic segmentation}
The system aggregates parsed document pieces into complete knowledge chunks by applying different content categories. Documents are segmented according to headers and sections, whereas source code is divided based on function and class boundaries. When individual chunks exceed an appropriate size, additional subdivision is performed to ensure that each unit remains within a length of 500 to 8,000 characters.
Markdown files follow hierarchical heading structures, while Python source files are segmented in accordance with intrinsic code organization patterns. Each segmentation strategy preserves contextual continuity. Oversized content units are automatically divided at semantically natural boundaries. This study avoids splitting sentences or code functions and uses parent-child relationships to link related blocks together. 
As a result of this procedure, the system produces 6,295 knowledge blocks for inclusion in the knowledge base. Each block is associated with metadata describing the file location, a concise content summary, and structural attributes required for subsequent processing.

\subsubsection{Entity relationship extraction and graph construction}

This stage performs entity extraction across 6,295 knowledge chunks and constructs the corresponding graph relationships. Three complementary extraction strategies are applied. The first approach relies on code analysis to detect Python classes, functions, and variables through regular expression patterns. The second approach employs domain dictionaries to recognize and match concepts unique to Devito, including elements such as \texttt{Grid}, \texttt{TimeFunction}, and \texttt{Operator}. The third approach uses SpaCy to carry out general entity recognition of technical terms and supplements.

The extractor creates five relationship types:
\begin{itemize}
\item \textbf{MENTIONS} links chunks to contained entities.
\item \textbf{CALLS} tracks function dependencies.
\item \textbf{INHERITS} maps class hierarchies.
\item \textbf{RELATED\_TO} connects similar content.
\item \textbf{PART\_OF} shows chunk structure.
\end{itemize}

The procedure generates a total of 12,793 nodes and 62,362 relationships. Entities derived from code encompass key Devito classes and essential computational functions. Conceptual entities represent topics such as finite difference schemes and boundary condition types. This structured mapping provides a foundational framework for subsequent community detection and database integration by encoding both the structural and semantic relationships inherent in the codebase.

\subsubsection{Embedding, community detection and graph optimization}

The project employs BGE-M3 (\texttt{BAAI/bge-m3}) as the primary embedding model to perform semantic similarity analysis. Each node within the graph is represented by an embedding vector generated by this model. In total, 32,416 nodes were processed, comprising 24,265 knowledge blocks and 8,151 extracted entities.
A similarity graph is constructed from the embedding vectors and analyzed using the Leiden algorithm. Multiple resolution settings (0.3, 0.5, 0.8, 1.0, 1.2) are evaluated to identify the most meaningful community structures. The algorithm detects 70 distinct semantic communities encompassing 6,230 nodes. Major communities include symbolic mathematics (690 nodes), seismic wave modeling (334 nodes), boundary conditions (223 nodes), and finite difference methods (175 nodes). This detailed community organization captures both the technical depth and the modular design characteristic of the Devito framework.

\begin{figure}[H]
\centering
\includegraphics[width=0.60\linewidth]{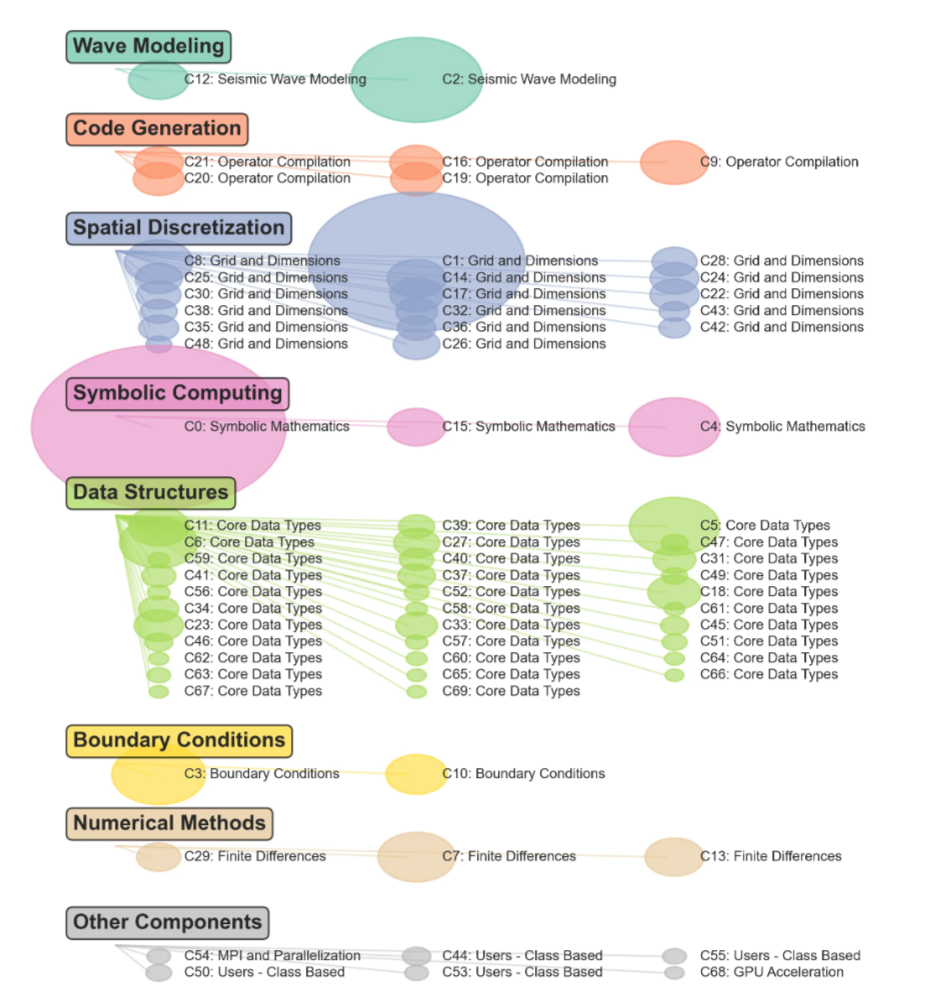}
\caption{Community hierarchy by theme categories.}
\label{fig:leiden_communities}
\end{figure}

Figure \ref{fig:leiden_communities} illustrates the hierarchical organization of semantic communities identified using the Leiden algorithm. The system automatically identifies multiple semantic clusters and organizes them into eight overarching technical domains based on thematic similarity. For instance, spatial discretization encompasses grid and dimension handling, fluctuation modeling covers seismic and acoustic propagation methods, code generation addresses operator compilation and optimization strategies, and data structures include core type definitions. This multi-level arrangement of knowledge enables efficient retrieval across varying levels of detail.
Graph optimization leverages a Top-$K$ nearest neighbor approach to reduce computational overhead. Each node maintains links only to its eight most similar neighbors, preventing the creation of an overly dense network. This method condenses hundreds of thousands of possible semantic relationships into 36,702 high-quality edges. The resulting sparsification enhances query performance by a factor of 540 for real-time retrieval tasks, while preserving essential semantic connections.

\subsubsection{Neo4j migration and index building}

The final stage of the pipeline involves migration to Neo4j. During this process, schema constraints and indexes are established. Unique constraints ensure the integrity of each node, while performance indexes are applied to community identifiers, chunk categories, and thematic classifications \ref{example_cypher_queries}.
To enhance import efficiency, batch processing is utilized. Nodes are imported in groups of 500 using parameterized queries, reducing both security risks and memory consumption. Parallel processing further accelerates the operation by simultaneously handling multiple data types.
Although executed only once, the full import completes in 17 seconds. The resulting database supports high-performance GraphRAG functionality, providing optimized query paths and robust semantic search capabilities.

\subsection{GraphRAG retrieval system design} 
\subsubsection{Fortran static analysis and query generation}

When Fortran code is provided as input, the system’s \texttt{FortranCodeAnalyzer} automatically detects features such as finite difference schemes, boundary condition types, and time-stepping approaches, converting them into structured query vectors. The analyzer extracts information on spatial dimensions, PDE classifications, and numerical method signatures, assigning confidence scores and complexity metrics to each identified code element.
The GraphRAG component employs a three-layer fixed strategy mapping architecture. Using results from static analysis of the Fortran source, the system generates three categories of characteristic queries and assigns them to predefined retrieval strategies:
\begin{itemize}
\item \textbf{Primary queries (Comprehensive strategy):} Core technical elements are mapped to a comprehensive retrieval strategy, ensuring high-fidelity coverage of essential conversion dependencies (e.g., “2D heat equation finite difference Devito implementation''). 
\item \textbf{Secondary queries (Fast strategy):} Queries targeting auxiliary implementation details are routed to a fast retrieval strategy, emphasizing quick response times (e.g., “boundary condition implementation'' or “grid initialization patterns''). 
\item \textbf{Concept queries (Deep strategy):} Queries requiring deeper conceptual insight are directed to a deep retrieval strategy, enabling the identification of implicit knowledge relationships (e.g., “mathematical equivalence verification'').
\end{itemize}

\subsubsection{Multi-modal parallel retrieval strategy}

\begin{figure*}[t] \centering \includegraphics[width=\linewidth,height=.70\textheight,keepaspectratio]{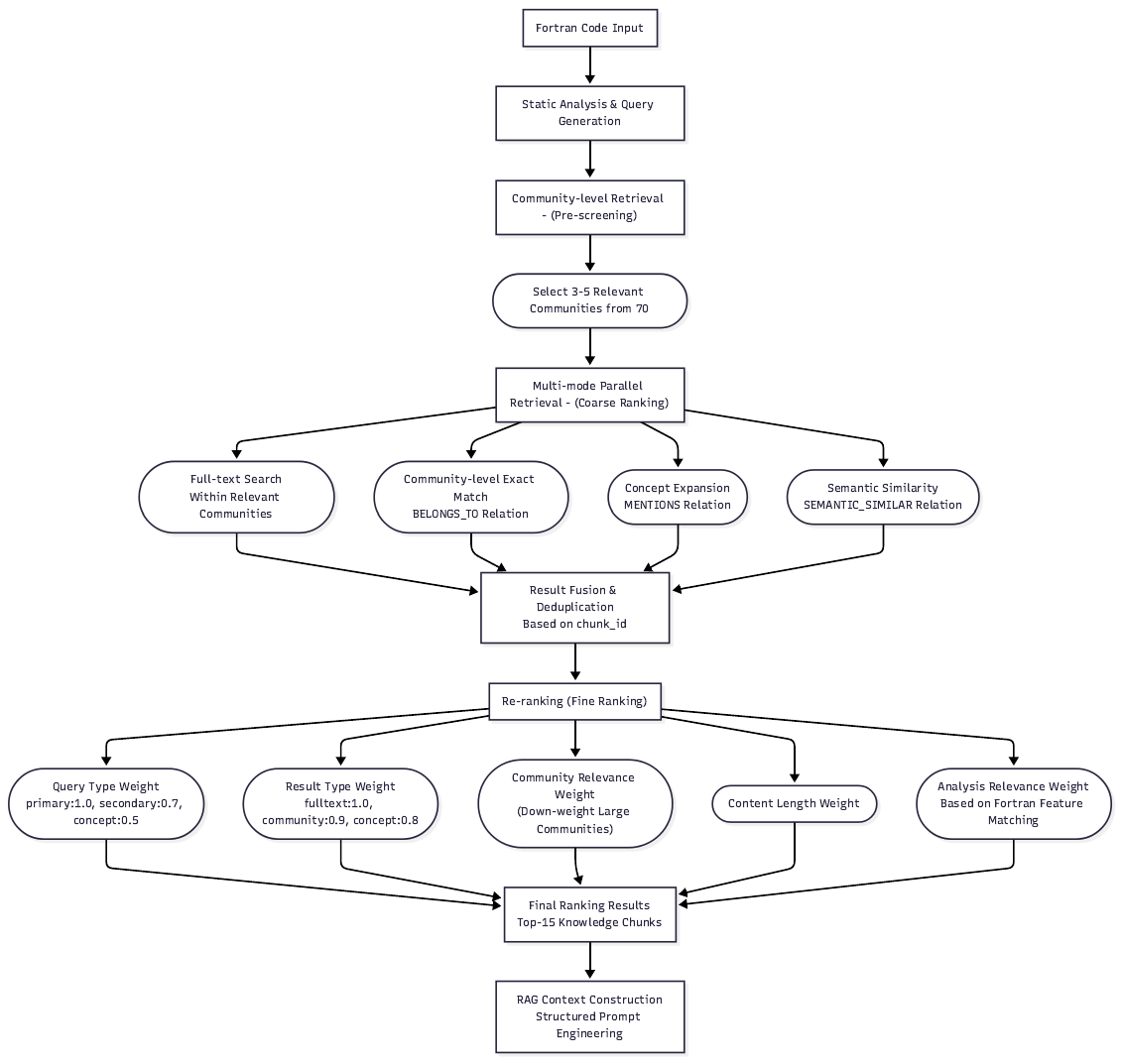} \caption{Retrieval-Augmented Generation (RAG) pipeline} 
\label{fig:rag_pipeline} 
\end{figure*}

The process initiates a multi-stage RAG retrieval pipeline that leverages the generated query types alongside the knowledge graph built in the preceding stage. Primary queries, following the comprehensive strategy, access information from the Neo4j graph database using four parallel retrieval mechanisms.
Figure \ref{fig:rag_pipeline} provides an overview of the full Retrieval-Augmented Generation (RAG) pipeline, summarizing its complete workflow.

The process begins with community-level retrieval: by comparing query keywords against the themes of 70 knowledge communities, the system rapidly selects the 3-5 most relevant communities, reducing the retrieval search space.
Next, four search modes are executed concurrently:
\begin{itemize}
\item \textbf{Full-text search:} Leveraging Neo4j’s pre-built full-text index and automatic TF-IDF scoring, this mode performs semantic matching within the selected communities.
\item \textbf{Precise community search:} Using string containment operations, this mode performs exact matching on knowledge chunk titles and content, constrained to specific communities via the \texttt{BELONGS\_TO} relationships.
\item \textbf{Concept expansion search:} Starting from identified concept nodes, related knowledge blocks are discovered through the \texttt{MENTIONS} relationship.
\item \textbf{Semantic similarity search:} Pre-computed cosine similarity scores (threshold 0.6) guide graph traversal, expanding results along the \texttt{SEMANTIC\_SIMILAR} relationships.
\end{itemize}
Each mode returns candidate results along with their original scores, providing multiple dimensions of evidence to support subsequent fusion and re-ranking.

\subsubsection{Result fusion and context construction}

Once the parallel retrieval phase completes the coarse ranking, the system advances to fine-grained ranking through advanced result merging and context assembly. The fusion process begins with deduplication using \texttt{chunk\_id}, ensuring that each knowledge chunk appears only once in the candidate set.
Composite scores are then recalculated using multi-factor weighting. Result-type weighting prioritizes full-text search (1.0), followed by community search (0.9), concept expansion (0.8), and semantic similarity (0.7). Community relevance weighting adjusts scores slightly to prevent dominance by large communities, while content-length weighting favors chunks of sufficient size, assigning full weight to items of 1,000 characters.
At the query-level, hierarchical weighted re-ranking is applied: primary queries carry a weight of 1.0, secondary queries 0.7, and concept queries 0.5.
An additional ``analysis relevance’’ factor dynamically adjusts scores between approximately 1.0 and 2.0, based on how well the features extracted from the Fortran code, such as equation types, numerical patterns, boundary conditions, and underlying mathematical concepts, align with the retrieved knowledge chunks.

The final phase generates a structured RAG context using layered prompt engineering. The prompt frames the LLM as a ``Fortran$\to$Devito conversion expert,'' specifying explicit objectives and quality standards, while the retrieved knowledge serves as the contextual knowledge base. Research indicates that structured prompt design enhances reasoning reliability, particularly when prompts incorporate role definition, constraints, and hierarchical task decomposition, as outlined in the Prompt Canvas framework \cite{promptcanvas2024} \ref{prompt_pseudo_code}.
To further improve conversion fidelity, the workflow incorporates a self-check and revision loop inspired by Chain-of-Thought (CoT) prompting. Studies have shown that CoT-based prompt engineering increases factual accuracy and reasoning depth in structured tasks \cite{cotprompting2024}. This mechanism enables the model to validate outputs against strict contracts, including Pydantic/JSON structures containing Devito code, rationale, mappings, and confidence scores, and iteratively refine them if requirements are unmet. Ultimately, the system selects the top 15 highest-scoring candidates, ensuring that only high-quality outputs guide the Devito conversion process.

\subsection{Agent workflow design}
\subsubsection{Concurrent workflow processing design}

During project development, a major bottleneck became apparent: sequential LLM API calls significantly reduced processing efficiency. Each code modification required 3-5 minutes for validation, which also impaired the responsiveness of interactive workflows. Testing of the \texttt{chat.ese} LLM API revealed full support for concurrent requests within the defined rate limits. Leveraging this capability, a concurrent task processing architecture was designed, enabling parallel execution of large-scale Fortran-to-Devito conversion tasks. The system balances concurrency control and resource management while remaining compliant with API constraints, dramatically improving throughput and responsiveness.

\subsubsection{I/O concurrency implementation}
The system employs Python’s asyncio framework to enable true I/O concurrency during interactions with the LLM API. Unlike traditional synchronous execution, where CPU resources remain idle while awaiting API responses, the asyncio event loop allows multiple tasks to progress simultaneously. When one workflow engine is paused waiting for an LLM response, the event loop immediately switches to other pending tasks, maximizing resource utilization in I/O-bound operations. Recent studies, such as \cite{Prompto2024}, have shown that asynchronous querying can substantially increase throughput when interfacing with LLM endpoints, and \cite{AsyncLM2024} highlights the benefits of non-blocking execution for complex LLM tasks. These findings confirm that asyncio-based concurrency is highly effective for large-scale LLM query processing and asynchronous function management.

\subsubsection{Task-level architecture}
The architecture adopts separate workflow agent instances assigned to individual conversion tasks. Each processing engine contains a fully self-sufficient pipeline. This design ensures complete isolation during task execution and integrates all required functional components, including source code transformation, output quality validation, and optimized layout generation.
\subsubsection{Concurrency control}
The system supports scalable execution ranging from two to eight concurrent large language model requests, achieving significant performance gains through the use of asyncio-based semaphore scheduling. Empirical evaluation using a representative benchmark dataset demonstrates notable acceleration. Relative to sequential execution, the configuration with two agents delivers a 2.15$\times$ throughput improvement, increasing productivity from 22.8 to 49.0 files per hour. Expanding to four agents further raises throughput to 131.2 files per hour, corresponding to a 5.75$\times$ speed increase.
The observed 5.75 times improvement in the four-agent scenario exceeds the nominal theoretical upper bound of four times. A parallel efficiency of 143.7\% highlights the substantial benefits derived from overlapping I/O operations during concurrent execution.

\subsection{Quality-driven iterative optimization}
\subsubsection{Iterative optimization method based on LangGraph}
This work develops a quality-based iterative optimization framework motivated by principles from reinforcement learning. Continuous enhancement of code conversion performance is achieved through decision routing in LangGraph combined with adaptive threshold calibration mechanisms.
This section defines navigation criteria derived from quantitative quality evaluation scores. Three decision thresholds are specified. An excellent level with a score of 8.5 identifies conversion outputs that satisfy verification requirements and can be released directly. An acceptable level with a score of 5.5 activates focused refinement procedures. A minimum level of 3 is established as the lowest permissible quality boundary.
Overall, the framework incorporates a four-path adaptive decision mechanism. High-quality outputs follow a direct pass route to maximize execution efficiency. Medium quality outputs undergo targeted refinement, where customized optimization guidance is generated according to identified defect categories. Low-quality outputs trigger a reconversion process, involving adjustments to conversion logic and processing strategies. When the predefined upper limit on large language model responses is reached, an exit procedure is applied, and the final handling strategy is selected based on the current quality assessment.
The introductory tutorial provided by LangChain demonstrates the use of the \texttt{get\_graph()} method for graph based visualization \cite{LangGraphReference}. In this study, the \texttt{get\_graph().draw\_mermaid\_png()} functionality of LangGraph is employed to generate workflow diagrams. State transition relationships are constructed using \texttt{StateGraph} \cite{ExsonJoseph2025} and are automatically rendered as intuitive flowchart representations.

\begin{figure}[H]
\centering
\includegraphics[width=0.7\linewidth,keepaspectratio]{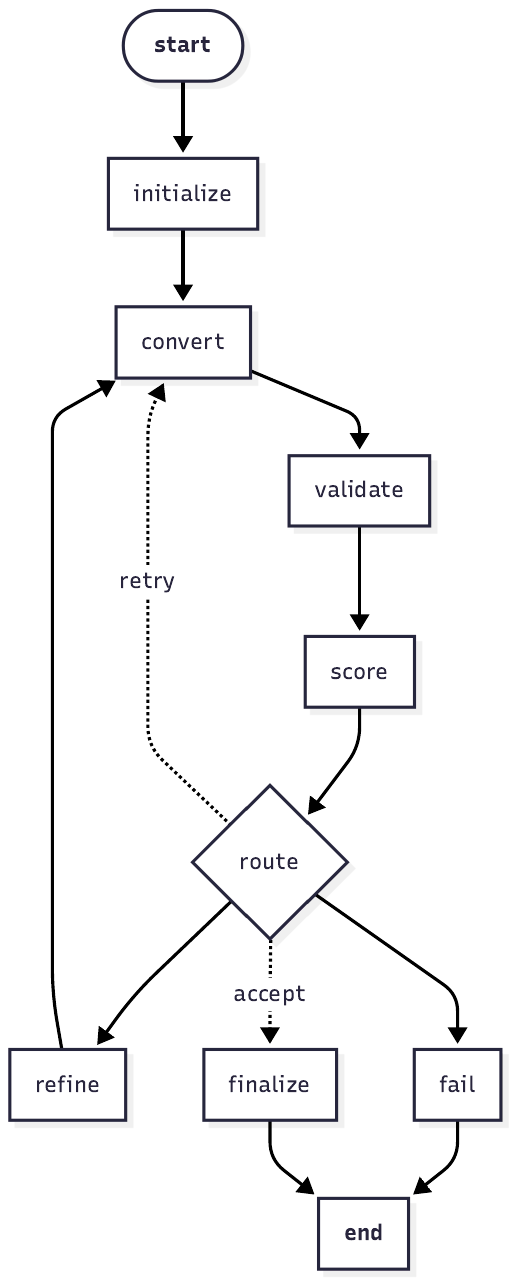}
\caption{LangGraph-Based optimization loop.}
\label{fig:langgraph_based_optimization_loop}
\end{figure}

As illustrated in Figure \ref{fig:langgraph_based_optimization_loop}, the project implements an iterative optimization procedure founded on the LangGraph framework, which follows a continuous enhancement loop. Each Fortran source file is processed through multiple optimization cycles. The workflow begins with an analysis of structural characteristics within the source code, followed by an initial conversion stage. The resulting output is then evaluated across multiple criteria by a dedicated quality verification module. Based on the resulting quality scores, the routing component determines the subsequent processing path \cite{Adarishanmukh2025}.

\subsubsection{Pydantic structured output constraints}
Through the strict type enforcement capabilities provided by Pydantic, large language models are constrained to produce structured JSON outputs that strictly adhere to a predefined schema. Each conversion output is required to contain complete Devito code, detailed explanations of conversion decisions, explicit component mapping relationships, confidence evaluations, and related metadata. This approach converts inherently unstructured model-generated text into a representation characterized by high structural fidelity and semantic coherence. As a result, the frequent formatting irregularities observed in conventional large language model-based code generation are effectively mitigated \cite{Liu2024WeNeedStructuredOutput, LangChainStructuredOutput}. The corresponding schema definition is presented in \ref{pydantic_schema}.

\subsubsection{Ruff code quality optimization}
Code formatting functionality provided by Ruff is incorporated into the Agent layer of the system. During the final stage of the processing workflow, Ruff is automatically executed to format the generated Devito source code. This step ensures compliance with Python PEP 8 conventions and widely accepted coding practices. The applied Ruff optimizations primarily focus on standardized code layout and systematic import arrangement, thereby improving both the readability and structural consistency of the produced code \cite{RuffFormatterDocs}.

\section{Evaluation and analysis}
\subsection{Retrieval quality assessment}
\subsubsection{Experimental design and methodology}
To assess retrieval effectiveness within a knowledge graph-based GraphRAG framework applied to Fortran to Devito code transformation tasks, a standardized retrieval quality evaluation protocol is constructed. The evaluation employs eleven benchmark queries designed to represent common information retrieval demands arising during the code conversion process.
The benchmark queries are grouped into three tiers of increasing complexity.
\begin{enumerate}
\item Basic pattern level queries, including concepts such as “finite difference methods’’ and “boundary conditions’’, are used to examine retrieval correctness for foundational numerical computing knowledge.
\item Intermediate complexity queries, such as “Devito time stepping implementation’’ and “2D heat equation solver optimization’’, aim to capture combined technical requirements involving multiple concepts.
\item Advanced technical queries, including “second-order spatial discretization stencil optimization’’ and “Devito compiler backend performance tuning’’, are intended to evaluate the system’s ability to retrieve highly specialized and expert-level information.
\end{enumerate}
These test cases reflect the progressive increase in complexity typically encountered in real-world code modernization pipelines \cite{Croft2010SearchEngines}.
Each query is paired with predefined ground truth references, consisting of expected topic categories, relevant keyword collections, and anticipated output types. Retrieval performance is evaluated using three widely adopted information retrieval metrics.
\begin{enumerate}
\item \textbf{Precision@5} quantifies the relevance accuracy of the top five retrieved items by computing the ratio of relevant results to the total number of returned entries \cite{Manning2008IntroductionIR}.
\item \textbf{Recall@5} measures the proportion of ground truth relevant items successfully retrieved, reflecting the completeness of the retrieval process \cite{PineconeRecall2023}.
\item \textbf{MRR Mean Reciprocal Rank} assesses the ranking position of the first relevant result, emphasizing the capability to prioritize the most pertinent information \cite{Voorhees1999TREC8}.
\end{enumerate}

\subsubsection{Experimental results and analysis}

The GraphRAG system demonstrates strong performance across all evaluated indicators, as reported in Table \ref{tab:retrieval_results}. The evaluation yields a Precision@5 value of 0.964 and a Recall@5 value of 0.930. These results indicate that the integration of knowledge graph community discovery, multi-path retrieval mechanisms, and semantic expansion strategies is effective for improving retrieval quality. In parallel, the mean query response time remains exceptionally low at 0.012 seconds. This efficiency is attributed to optimized indexing provided by Neo4j, combined with graph sparsification techniques and concurrent query execution. Collectively, these findings demonstrate that the system delivers both high-accuracy information retrieval and real-time responsiveness.

\begin{table}[H]
\centering
\caption{GraphRAG system retrieval quality evaluation results.}
\label{tab:retrieval_results}
\begin{tabular}{lc}
\toprule
\textbf{Metric} & \textbf{Value} \\
\midrule
Precision@5 & 0.964 \\
Recall@5 & 0.930 \\
MRR & 1.000 \\
Avg. Response Time & 0.012s \\
\bottomrule
\end{tabular}
\end{table}

The reported MRR values “look too good’’. This outcome arises because each query is associated with a \emph{set} of valid reference answers rather than a single unique target. When the top-ranked retrieval result belongs to the predefined ground truth set, MRR=1. Although this behavior reflects strong ranking consistency, it also limits the ability of MRR to differentiate retrieval quality under the current evaluation configuration.
\subsubsection{Multi-modal retrieval strategy analysis}
This experiment examines the three retrieval modes implemented within the GraphRAG system.
\begin{enumerate}
\item \textbf{Fast Strategy}: This mode prioritizes response latency reduction and relies exclusively on direct matching retrieval supported by full-text indexing in Neo4j. The technical realization includes straightforward full-text index queries, basic keyword matching operations, and simplified result ranking procedures.
\item \textbf{Comprehensive Strategy}: This mode deploys a complete four-stage retrieval pipeline. The process begins with community level search to restrict the candidate scope. It then performs parallel full-text retrieval and precise intra-community search. Subsequently, concept expansion is applied, followed by semantic similarity-based retrieval to refine the final results.
\item \textbf{Deep Strategy}: This mode extends the Comprehensive approach by introducing an additional semantic expansion phase. Deeper relational knowledge is uncovered through two-hop connections within the graph structure. The implementation leverages path-based queries supported by Neo4j using the Cypher query language, enabling exploration of indirect semantic associations.

\begin{lstlisting}[language=SQL]
MATCH (seed:Chunk)-[:SEMANTIC_SIMILAR*2]-(distant:Chunk)
\end{lstlisting}

This methodology is enabled by the efficient path-based query capabilities of the Cypher language for graph database retrieval tasks \cite{Robinson2015}.
\end{enumerate}

Each retrieval strategy independently processed an identical set of ten benchmark queries, spanning multiple levels of complexity and diverse query categories, while employing a unified set of evaluation criteria to ensure comparability.
The experimental design applies a multidimensional assessment framework comprising five evaluation dimensions. A detailed comparative analysis of the performance achieved by the four retrieval strategies is presented in Table \ref{tab:retrieval_performance}.

\begin{table*}[t]
\centering
\caption{Multi-modal retrieval strategy performance comparison analysis.}
\label{tab:retrieval_performance}
\resizebox{\textwidth}{!}{%
\begin{tabular}{lccccc}
\toprule
\textbf{Strategy Name} & \textbf{Avg Response Time (s)} & \textbf{Avg Precision@5} & \textbf{Avg Recall@5} & \textbf{Avg Diversity} & \textbf{Success Rate} \\
\midrule
Fast & 0.002 & 0.960 & 0.625 & 0.700 & 1.000 \\
Comprehensive & 0.017 & 0.980 & 0.668 & 0.595 & 1.000 \\
Deep & 0.018 & 0.980 & 0.668 & 0.595 & 1.000 \\
Hybrid & 0.013 & 0.980 & 0.654 & 0.635 & 1.000 \\
\bottomrule
\end{tabular}%
}
\end{table*}

The Fast strategy excelled in response speed, achieving the shortest average query time of 0.002 seconds, but exhibited a relatively lower recall of 0.625. In contrast, the Comprehensive strategy achieved an exceptional precision of 0.980 and a recall of 0.668, confirming the effectiveness of its multi-stage retrieval pipeline. Overall, this configuration provided the most favorable compromise between retrieval efficiency and output quality.
\subsection{Code translation quality assessment}
\subsubsection{API consistency and error mitigation}
During the initial deployment of the system, large language models occasionally produced outdated or non-existent APIs, such as \texttt{u.dx.backward} and \texttt{u.data.initialize()}. This behavior arises from the presence of early version examples and outdated Devito parameters in LLM training datasets. Additional errors were introduced through analogies with similar frameworks and flawed reasoning patterns. Such occurrences align with prior observations that LLMs frequently hallucinate or misuse APIs \cite{Chen2021codex, Amershi2019SEforML}.

To address this issue, a version locked retrieval augmented generation framework was implemented, governed by rule-based controls and informed by studies on LLM code reliability \cite{He2023}. The underlying knowledge base contains only verified Devito examples \cite{DevitoTutorials}, ensuring that retrieved outputs are executable. The system enforces explicit denylist and allowlist rules, along with equivalent substitution mappings. Additional safeguards include Preflight structural verification to ensure assignable left-hand values, consistent equation dimensions, and valid subdomains and subdimensions; Python syntax validation through abstract syntax tree parsing; and API compliance linting \cite{Louboutin2019DevitoDSL, Luporini2020Devito}.
To demonstrate the improvement in code correctness provided by the RAG framework compared to native LLM outputs, several representative error cases have been compiled in Appendix~A. These include instances of fake APIs and ``pretend'' Devito but calculate with NumPy. (see \ref{appendix:rag_vs_llm}).

\subsubsection{Multi-dimensional quality validation}
A critical component for ensuring successful translation from Fortran to Devito is the assessment of code conversion quality. This study establishes a quality verification framework that integrates the G-Eval evaluation methodology with conventional static analysis techniques. The GPT OSS 120B model was employed as the large language model evaluator within the open-source environment.
The static quality assessment model in this framework evaluates code along five key dimensions:
\begin{itemize}
\item \textbf{Execution success}: evaluates whether the converted code can run without errors, accounting for runtime exception handling, completeness of imports, and syntactic correctness.
\item \textbf{Structural integrity}: assesses the presence and correctness of Devito core components, including Grid definitions, Function instantiations, Equation formulations, and Operator construction.
\item \textbf{API compliance}: measures adherence to Devito API usage best practices through pattern recognition, parameter configuration validation, and verification of API call structures.
\item \textbf{Parameter consistency}: examines the accurate translation of time, grid, and physical parameters, ensuring boundary conditions, dimension alignment, and numerical consistency are maintained.
\item \textbf{Conversion fidelity}: determines the precision of translating mathematical models and physical equations, including correct equation type mapping, preservation of differential operators, and maintenance of physical semantics to guarantee the correctness of scientific computations.
\end{itemize}

The comprehensive quality score is calculated using weighted averaging:

\begin{equation}
\label{eq:quality_score}
\text{Quality\_Score} = \sum (w_i \times s_i)
\end{equation}

In this expression. $w_i$ is the weight of the $i$-th dimension, and $s_i$ represents the standardized score of the $i$-th dimension.

G-Eval is an automated evaluation framework that leverages large language models to assess generated content through structured prompt design and sequential reasoning. Following OpenAI’s official RAG system evaluation guidelines, the GPT OSS 120B open-source model was employed as the evaluation engine to complement the traditional five-dimensional static analysis framework \cite{g_eval_2023, openai-rag-eval}. This approach enables the handling of complex assessment tasks that are difficult to quantify using conventional metrics, including code logic, algorithmic elegance, and adherence of the code to the original specifications. Carefully designed instructions and illustrative examples guide the large language model to evaluate outputs according to predefined criteria.
To ensure the objectivity of the evaluation, a model distinct from the one used for code generation was applied, following established practices in the LLM-as-a-Judge domain. Prior research indicates that using the same model for both generation and evaluation introduces substantial self-enhancement bias, with models tending to assign inflated scores to their own outputs. To mitigate this effect, the GPT OSS 120B model functions solely as a separate evaluation engine, independent of the code conversion model.
The evaluation procedure applies multi-criteria scoring across four dimensions: execution success (30\%, code runs without errors), code structure (25\%, Devito core components completeness), mathematical logic (25\%, solution validity within the same PDE framework), and API usage (20\%, appropriateness of Devito API calls). Both the original Fortran code and the converted Devito code are provided as joint context input. To eliminate variability caused by stochastic outputs and ensure reproducibility, the model’s temperature is set to zero \cite{blackwell2024reproducible}. Additionally, the model incorporates an implicit chain-of-thought reasoning process for internal step-by-step evaluation and weighting. The output consists solely of a final score accompanied by a brief justification.

The overall score is calculated considering both the LLM and the traditional static method:
\begin{equation}
\label{eq:final_score}
\mathrm{Final\_Score} = \mathrm{Traditional\_Score} \times (1 - \lambda) 
+ \mathrm{LLM\_Score} \times \lambda, \quad \lambda = 0.5
\end{equation}

Table \ref{tab:g_eval_full_single_tt} presents the evaluation results for 13 Devito finite difference test cases covering varying difficulty levels and application domains. The assessment indicates strong system performance, with a Grade-A success rate of 76.9\%. All conversions achieved perfect scores of 1.00 in the three critical dimensions: execution, structural integrity, and API compliance. Furthermore, the scores generated by the static assessment framework show a positive correlation with the G-Eval model judgments, demonstrating convergent validity and supporting the reliability of the proposed evaluation methodology.

\begin{table*}[t]
\centering
\caption{Unified quality validation results per case.}
\label{tab:g_eval_full_single_tt}
\resizebox{\textwidth}{!}{%
\begin{tabular}{lllllllllll}
\toprule
Case & Final & Grade & Confidence & Duration (s) & Execution & Structure & API & Parameters & Conv. Fidelity & LLM Judge \\
\midrule
\texttt{acoustic\_wave\_2d} & 0.941 & A & 0.750 & 36.090 & 1.000 & 1.000 & 1.000 & 1.000 & 0.820 & 0.900 \\
\texttt{advection\_simple} & 0.892 & A & 0.900 & 33.290 & 1.000 & 1.000 & 1.000 & 1.000 & 0.840 & 0.800 \\
\texttt{advection\_upwind} & 0.780 & B & 0.750 & 36.410 & 1.000 & 1.000 & 1.000 & 0.950 & 0.640 & 0.600 \\
\texttt{crank\_nicolson\_heat} & 0.930 & A & 0.750 & 15.220 & 1.000 & 1.000 & 1.000 & 1.000 & 0.600 & 0.900 \\
\texttt{diffusion\_3d} & 0.883 & A & 0.750 & 9.160 & 1.000 & 1.000 & 1.000 & 0.950 & 0.700 & 0.800 \\
\texttt{heat\_1d\_simple} & 0.842 & A & 0.900 & 11.150 & 1.000 & 1.000 & 1.000 & 1.000 & 0.840 & 0.700 \\
\texttt{heat\_equation\_2d} & 0.842 & A & 0.950 & 15.860 & 1.000 & 1.000 & 1.000 & 1.000 & 0.840 & 0.700 \\
\texttt{laplace\_solver} & 0.877 & A & 0.750 & 9.200 & 1.000 & 1.000 & 1.000 & 0.800 & 0.740 & 0.800 \\
\texttt{legacy\_advection} & 0.886 & A & 0.900 & 10.410 & 1.000 & 1.000 & 1.000 & 0.750 & 0.960 & 0.800 \\
\texttt{poisson\_jacobi} & 0.827 & A & 0.750 & 8.610 & 1.000 & 1.000 & 1.000 & 0.800 & 0.740 & 0.700 \\
\texttt{poisson\_simple} & 0.877 & A & 0.750 & 8.570 & 1.000 & 1.000 & 1.000 & 0.800 & 0.740 & 0.800 \\
\texttt{wave\_1d\_simple} & 0.729 & B & 0.750 & 15.300 & 1.000 & 1.000 & 1.000 & 0.900 & 0.680 & 0.500 \\
\texttt{wave\_equation\_1d} & 0.729 & B & 0.750 & 15.030 & 1.000 & 1.000 & 1.000 & 0.900 & 0.680 & 0.500 \\
\bottomrule
\end{tabular}
}
\end{table*}

The result is illustrated by the radar chart provided in \ref{appendix:radar_chart}.

\section{Discussion}
\subsection{Current limitations}
Although the system is capable of collecting and storing comprehensive conversion history data, its capabilities for advanced data mining and pattern discovery remain limited.
Additionally, the existing quality evaluation framework relies on fixed, predefined thresholds, such as \texttt{excellent\_threshold=0.85} and \texttt{acceptable\_threshold=0.55}. These criteria are not yet capable of automatic adjustment based on code type, complexity, or historical performance. Consequently, the system’s adaptability across diverse conversion scenarios is constrained by this static configuration.

\subsection{Future research directions}
Future developments will extend beyond using a single open-source large language model as the evaluation engine. Specialized, compact evaluator models will be trained or fine-tuned on historical conversion traces, while reinforcement learning techniques will dynamically adjust routing policies and quality thresholds, enabling a self-improving agent.
An agent-level memory system will also be introduced, storing episodic logs of previous conversions and semantic summaries of recurring error patterns. This memory allows the retrieval of prior fixes and the reuse of successful strategies across tasks. By integrating adaptive learning, this approach enhances both the reliability and efficiency of code modernization workflows.

\section{Conclusion}

This study presents an AI agent system designed to translate legacy Fortran finite difference code into the Devito framework. The system integrates Retrieval-Augmented Generation with open-source large language models within a multi-stage iterative workflow structured on the LangGraph architecture.
The agent utilizes a Devito knowledge graph implemented in Neo4j, incorporates GraphRAG for retrieval, and applies quality-driven iterative refinement with enforced structured output constraints. Conversion quality is ensured through a validation framework that combines static code analysis with LLM-based evaluation.
The resulting framework provides an agent-driven solution for automated legacy code translation in scientific computing, achieving a balance between accuracy and computational efficiency.

\appendices

\section{RAG-Governed Devito vs. Vanilla LLM}
\label{appendix:rag_vs_llm}

We have listed 3 categories of high-frequency errors:
\begin{itemize}
\item Hallucinated or mis-ported Devito APIs (e.g., \texttt{u.dx.backward}).
\item Devito integration in name only (imports Devito but performs updates/loops in NumPy).
\item Boundary-condition misuse (e.g., post-hoc patching or \texttt{bc=} on \texttt{Operator}).
\end{itemize}

\subsection*{A.1 Fabricated Derivative API (u.dx.backward)}
\textbf{Tag:} Treating derivatives as chained attributes; misinterpreting central difference as upwind.

\textbf{Incorrect (minimal example):}

\begin{lstlisting}[language=Python]
from devito import Grid, TimeFunction, Eq, solve, Operator, Constant

nx, Lx = 200, 1.0
grid = Grid(shape=(nx,), extent=(Lx,))
x, = grid.dimensions
u = TimeFunction(name='u', grid=grid, time_order=1, space_order=1)
c = Constant(name='c', value=1.0)

# Fabricated API: u.dx.backward (not supported in Devito)
eq = Eq(u.dt, -c * u.dx.backward)
stencil = solve(eq, u.forward)
op = Operator(Eq(u.forward, stencil))
\end{lstlisting}

\textbf{Correct:}
\begin{lstlisting}[language=Python]
from devito import Grid, TimeFunction, Eq, solve, Operator, Constant, first_derivative

nx, Lx = 200, 1.0
grid = Grid(shape=(nx,), extent=(Lx,))
x, = grid.dimensions
u = TimeFunction(name='u', grid=grid, time_order=1, space_order=1)
c = Constant(name='c', value=1.0)

# First-order upwind: explicit side (c>0 -> left; c<0 -> right)
du_dx = first_derivative(u, dim=x, side='left')
eq = Eq(u.dt, -c * du_dx)
stencil = solve(eq, u.forward)
op = Operator([Eq(u.forward, stencil)])
\end{lstlisting}

\subsection*{A.2 Pseudo Integration: Devito Wrapper, NumPy Core}
\textbf{Tag:} Devito package imported but not used; all updates performed in NumPy loops; boundary conditions patched after execution.

\textbf{Incorrect (pseudo integration, Devito unused in core):}
\begin{lstlisting}[language=Python]
from devito import Grid, TimeFunction, Eq, Operator # imported but never used
import numpy as np

nx, Lx = 100, 1.0
dx = Lx / nx
dt, c = 0.005, 1.0

# Allocate manual NumPy array (ignores Devito TimeFunction)
u = np.zeros((501, nx), dtype=np.float32)
xcoords = np.linspace(0.0, Lx - dx, nx, dtype=np.float32)
u[0, (xcoords > 0.2) & (xcoords < 0.4)] = 1.0

# Manual time stepping loop
for t in range(500):
    u_next = u[t].copy()
    for i in range(1, nx):
        du_dx = (u[t, i] - u[t, i-1]) / dx
        u_next[i] = u[t, i] - dt * c * du_dx
    # Incorrect periodic boundary: patched after update
    u_next[0] = u_next[-1]
    u[t+1] = u_next
\end{lstlisting}

\textbf{Correct (upwind + periodic BC encoded symbolically in Devito):}
\begin{lstlisting}[language=Python]
from devito import Grid, TimeFunction, Eq, Operator, Constant, first_derivative
import numpy as np

nx, Lx = 100, 1.0
dx = Lx / nx
dt_val, c_val = 0.005, 1.0

grid = Grid(shape=(nx,), extent=(Lx,))
x, = grid.dimensions
u = TimeFunction(name='u', grid=grid, time_order=1, space_order=1)
c = Constant(name='c', value=c_val)

xcoords = np.linspace(0.0, Lx - dx, nx, dtype=np.float32)
u.data[0, :] = ((xcoords > 0.2) & (xcoords < 0.4)).astype(np.float32)

du_dx = first_derivative(u, dim=x, side='left') # c>0 -> left
core = Eq(u.forward, u - dt_val * c * du_dx)

# Periodic BC: endpoints equal
left_bc = Eq(u.forward.subs({x: x.symbolic_min}), u.forward.subs({x: x.symbolic_max}))
right_bc = Eq(u.forward.subs({x: x.symbolic_max}), u.forward.subs({x: x.symbolic_min}))

op = Operator([core, left_bc, right_bc])
op.apply(time_M=499)
\end{lstlisting}

\subsection*{A.3 Boundary Condition Misuse + Variable/Index Errors}
\textbf{Tag:} Using bc= as Operator arg; SubDomain misuse; 1-based indexing (Fortran habit); variable name typos.

\textbf{Incorrect (minimal example):}
\begin{lstlisting}[language=Python]
from devito import *
import numpy as np

grid = Grid(shape=(101,), extent=(1.0,))
u = TimeFunction(name='u', grid=grid, space_order=1)
dt, c, nx, nsteps = 1e-4, 1.0, 100, 500

# 1-based indexing + variable name typo (nu_data not defined)
u_data = np.zeros((nx + 1), dtype=np.float64)
for i in range(1, nx + 1):
    x = (i - 1) * 0.01
    if 0.2 <= x <= 0.4:
        nu_data[i] = 1.0
    u.data[0][1:] = nu_data # typo + offset assignment

# Operator(bc=...) not valid API; SubDomain one-liner incorrect
bcs = [SubDomain('x==0', {'u': 0}), SubDomain('x==1', {'u': 0})]
eq = Eq(u.dt, -c * u.dx) # also uses central diff instead of upwind
op = Operator(eq, bc=bcs) # invalid/legacy API usage
op.apply(time_M=nsteps-1, dt=dt)
\end{lstlisting}

\textbf{Correct (minimal fix: 0-based indexing; BC as equations):}
\begin{lstlisting}[language=Python]
from devito import Grid, TimeFunction, Eq, Operator, Constant, first_derivative
import numpy as np

nx, Lx = 100, 1.0
grid = Grid(shape=(nx,), extent=(Lx,))
x, = grid.dimensions
u = TimeFunction(name='u', grid=grid, time_order=1, space_order=1)

initial = np.zeros(nx, dtype=np.float64)
xs = np.arange(nx) * (Lx / nx)
initial[(xs >= 0.2) & (xs <= 0.4)] = 1.0
u.data[0, :] = initial

dt_val, c = 1e-4, 1.0
core = Eq(u.forward, u - dt_val * c * u.dx) # use first_derivative(...) for upwind

left_bc = Eq(u.forward.subs({x: x.symbolic_min}), 0.0)
right_bc = Eq(u.forward.subs({x: x.symbolic_max}), 0.0)

op = Operator([core, left_bc, right_bc])
op.apply(time_M=499)
\end{lstlisting}

\section{Radar chart}
\label{appendix:radar_chart}
\begin{figure}[H]
\centering
\includegraphics[width=0.7\linewidth]{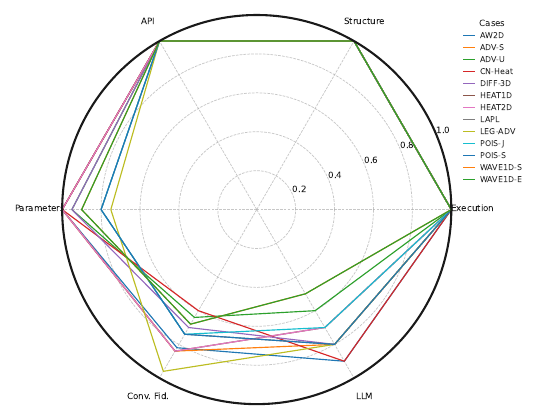}
\caption{Radar chart of evaluation results across 13 test cases}
\label{fig:g_eval_radar}
\end{figure}

\section{Pydantic schema}
\label{pydantic_schema}
\textbf{Pydantic schema for structured Fortran-to-Devito output:}
\begin{lstlisting}[language=Python]
Schema FortranToDevitoConversion:
devito_code: string
conversion_summary: string
key_decisions: list of {decision_type, rationale}
devito_components: list of {component, purpose}
equation_type: {parabolic, hyperbolic, elliptic}
spatial_dimensions: integer (1-3)
time_dependent: boolean
conversion_confidence: float (0.0-1.0)
validation: {execution_success, structure, api_compliance,
parameters, fidelity}
usage_notes: list of strings
optimization_hints: list of strings
\end{lstlisting}

\section{Prompt pseudo-code}
\label{prompt_pseudo_code}
\textbf{Prompt construction logic (simplified pseudo-code):}
\begin{lstlisting}[language=Python]
Function BuildConversionPrompt(fortran_code, rag_results):

Extract analysis info: equation_type, dimensions, complexity
Extract reference examples from rag_results

Initialize prompt_parts

Append task definition and conversion requirements
Append standard Devito workflow pattern

If analysis info exists:
Append problem type, dimensions, complexity

If reference examples exist:
Append retrieved examples

Append source Fortran code
Append implementation guidelines
Append required JSON output format (schema)

Return full structured prompt
\end{lstlisting}

\section{Example Cypher Queries}
\label{example_cypher_queries}

Here is a list of several basic Cypher queries for Neo4j.

\begin{lstlisting}[language=SQL]
MATCH (n) 
RETURN labels(n)[0] AS NodeType, count(*) AS Count 
ORDER BY Count DESC
\end{lstlisting}

This query counts the number of nodes by type, providing a global overview of the database composition.

\begin{lstlisting}[language=SQL]
MATCH (com:Community) 
RETURN com.id, com.theme, com.size 
ORDER BY com.size DESC 
LIMIT 10
\end{lstlisting}

This query lists the ten largest semantic communities.

\begin{lstlisting}[language=SQL]
MATCH (c:Chunk)-[:BELONGS_TO]->(com:Community)
RETURN com.theme, count(c) AS member_count
ORDER BY member_count DESC
LIMIT 10
\end{lstlisting}

\begin{figure}[H]
\centering
\includegraphics[width=0.9\linewidth]{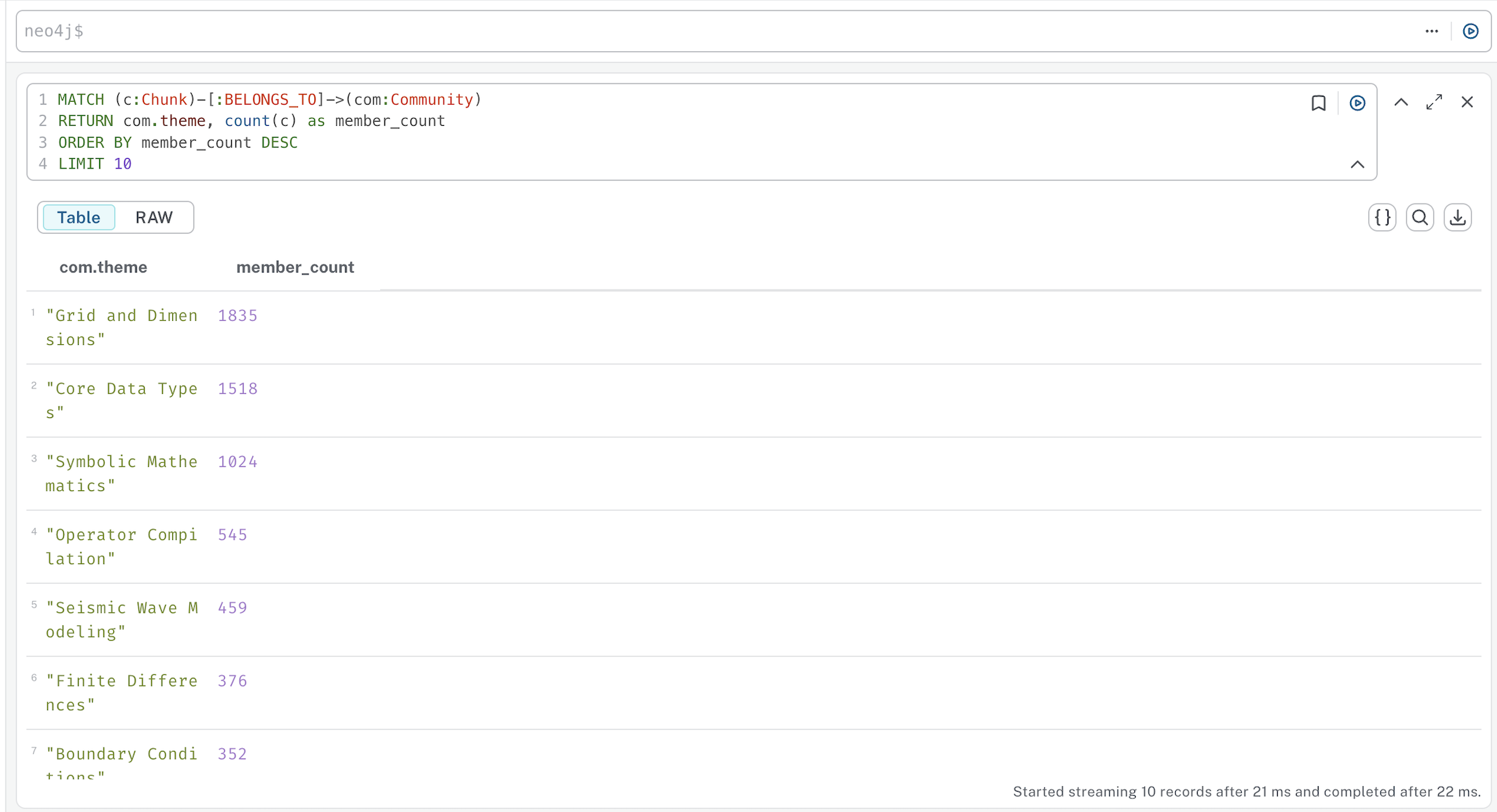}
\caption{cypher example}
\label{fig:cypher}
\end{figure}

This query verifies the assignment of knowledge chunks to communities by counting their membership distribution. Figure~\ref{fig:cypher} shows the top communities and their sizes returned by this query.

 \label{appendix:cypher_example}

\begin{lstlisting}[language=SQL]
CALL db.index.fulltext.queryNodes("chunk_content_index", "finite difference") 
YIELD node, score 
RETURN node.title, score 
LIMIT 5
\end{lstlisting}

This query performs a semantic search using the full-text index, retrieving chunks most relevant to the keyword \textit{finite difference}.

 \bibliographystyle{plain}
 \bibliography{references}

@article{promptcanvas2024,
  author = {Amatriain, Xavier},
  year = {2024},
  title = {Prompt Design and Engineering: Introduction and Advanced Methods},
  journal = {arXiv preprint arXiv:2401.14423},
  note = {Submitted on 24 January 2024; revised 5 May 2024},
  doi = {10.48550/arXiv.2401.14423}
}

@article{cotprompting2024,
  author = {Hewing, Michael and Leinhos, Vincent},
  year = {2024},
  title = {The Prompt Canvas: A Literature-based Practitioner Guide for Creating Effective Prompts in Large Language Models},
  journal = {arXiv preprint arXiv:2412.05127},
  note = {Submitted on 6 December 2024},
  doi = {10.48550/arXiv.2412.05127}
}

@article{prompto2024,
  author = {Chan, Ryan Sze-Yin and Nanni, Federico and Williams, Angus R. and Brown, Edwin and Burke-Moore, Liam and Chapman, Ed and Onslow, Kate and Sippy, Tvesha and Bright, Jonathan and Gabasova, Evelina},
  year = {2024},
  title = {Prompto: An open source library for asynchronous querying of LLM endpoints},
  journal = {arXiv preprint arXiv:2408.11847},
  note = {Submitted on 12 August 2024; revised 16 December 2024},
  doi = {10.48550/arXiv.2408.11847},
  url = {https://arxiv.org/abs/2408.11847}
}

@article{AsyncLM2024,
  author = {In Gim and Seung-seob Lee and Lin Zhong},
  title = {Asynchronous LLM Function Calling},
  journal = {arXiv preprint arXiv:2412.07017},
  year = {2024},
  note = {Published on 9 December 2024},
  doi = {10.48550/arXiv.2412.07017},
  url = {https://arxiv.org/abs/2412.07017}
}

@misc{LangGraphReference,
  author       = {{LangGraph Project}},
  year         = {2025},
  title        = {LangGraph Graphs Reference},
  howpublished = {Online},
  note         = {Available from: \url{https://langchain-ai.github.io/langgraph/reference/graphs/} [Accessed 18 August 2025]}
}

@misc{ExsonJoseph2025,
  author       = {Exson, J.},
  year         = {2025},
  title        = {Langgraph visualization with get\_graph},
  howpublished = {Online},
  note         = {Medium. Available from: \url{https://medium.com/@josephamyexson/langgraph-visualization-with-get-graph-ffa45366d6cb} [Accessed 18 August 2025]}
}

@misc{Adarishanmukh2025,
  author       = {Adarishanmukh},
  year         = {2025},
  title        = {Langraph Conditional WorkFlow},
  howpublished = {Online},
  note         = {Medium. Available from: \url{https://medium.com/@adarishanmukh15501/langraph-c61c8fcaac8f} [Accessed 18 August 2025]}
}

@inproceedings{Liu2024WeNeedStructuredOutput,
  author    = {Liu, Michael Xieyang and Liu, Frederick and Fiannaca, Alexander J. and Koo, Terry and Dixon, Lucas and Terry, Michael and Cai, Carrie J.},
  title     = {“We Need Structured Output”: Towards User-centered Constraints on Large Language Model Output},
  booktitle = {Extended Abstracts of the CHI Conference on Human Factors in Computing Systems (CHI EA ’24), 11–16 May 2024, Honolulu, HI, USA},
  year      = {2024},
  publisher = {ACM},
  doi       = {10.1145/3613905.3650756}
}

@MISC{LangChainStructuredOutput,
  title        = {Structured outputs},
  author       = {{LangChain Documentation}},
  howpublished = {Online},
  year         = {2025},
  note         = {LangChain Documentation. Available from: \url{https://python.langchain.com/docs/concepts/structured_outputs/} [Accessed 18 August 2025]}
}

@MISC{RuffFormatterDocs,
  title        = {The Ruff Formatter},
  author       = {{Ruff Documentation}},
  howpublished = {Online},
  year         = {2025},
  note         = {Astral Documentation. Available from: \url{https://docs.astral.sh/ruff/formatter/} [Accessed 18 August 2025]}
}

@BOOK{Croft2010SearchEngines,
  author    = {Croft, W. Bruce and Metzler, Donald and Strohman, Trevor},
  year      = {2010},
  title     = {Search engines: information retrieval in practice},
  publisher = {Addison-Wesley},
  address   = {Boston, MA},
  isbn      = {9780136072249}
}

@BOOK{Manning2008IntroductionIR,
  author    = {Manning, Christopher D. and Raghavan, Prabhakar and Schütze, Hinrich},
  year      = {2008},
  title     = {Introduction to information retrieval},
  publisher = {Cambridge University Press},
  address   = {Cambridge, UK},
  isbn      = {9780521865715}
}

@INPROCEEDINGS{Voorhees1999TREC8,
  author    = {Voorhees, Ellen M.},
  year      = {1999},
  title     = {The TREC-8 question answering track report},
  booktitle = {Proceedings of the Eighth Text REtrieval Conference (TREC-8), November 1999, Gaithersburg, MD, USA},
  publisher = {National Institute of Standards and Technology (NIST)},
  address   = {Gaithersburg, MD},
  pages     = {77--82},
  series    = {NIST Special Publication},
  volume    = {500-246}
}

@MISC{PineconeRecall2023,
  author       = {{Pinecone}},
  year         = {2023},
  title        = {Offline evaluation for ranking models: precision and recall at K},
  howpublished = {Online},
  note         = {Available from: \url{https://www.pinecone.io/learn/offline-evaluation/} [Accessed 18 August 2025]}
}

@book{Robinson2015,
  author    = {Robinson, Ian and Webber, Jim and Eifrem, Emil},
  year      = {2015},
  title     = {Graph databases},
  edition   = {2nd},
  publisher = {O'Reilly Media},
  address   = {Sebastopol, CA},
  isbn      = {978-1-491-93516-5}
}

@inproceedings{He2023,
  author    = {He, Jingxuan and Vechev, Martin},
  year      = {2023},
  title     = {Large language models for code: security hardening and adversarial testing},
  booktitle = {Proceedings of the 2023 ACM SIGSAC Conference on Computer and Communications Security (CCS '23), November 2023, Copenhagen, Denmark},
  publisher = {ACM},
  address   = {New York, NY},
  pages     = {1865--1879},
  doi       = {10.1145/3576915.3623175}
}

@article{Chen2021codex,
  author  = {Chen, Mark and Tworek, Jerry and Jun, Heewoo and Yuan, Qiming and de Oliveira Pinto, Henrique Ponde and Kaplan, Jared and Edwards, Harri and Burda, Yuri and Joseph, Nicholas and Brockman, Greg and Ray, Alex and Puri, Raul and Krueger, Gretchen and Petrov, Michael and Khlaaf, Heidy and Sastry, Girish and Mishkin, Pamela and Chan, Brooke and Gray, Scott and Ryder, Nick and Pavlov, Mikhail and Power, Alethea and Kaiser, Lukasz and Bavarian, Mohammad and Winter, Clemens and Tillet, Philippe and Such, Felipe Petroski and Cummings, Dave and Plappert, Matthias and Chantzis, Fotios and Barnes, Elizabeth and Herbert-Voss, Ariel and Guss, William Hebgen and Nichol, Alex and Paino, Alex and Tezak, Nikolas and Tang, Jie and Babuschkin, Igor and Balaji, Suchir and Jain, Shantanu and Saunders, William and Hesse, Christopher and Carr, Andrew N. and Leike, Jan and Achiam, Josh and Misra, Vedant and Morikawa, Evan and Radford, Alec and Knight, Matthew and Brundage, Miles and Murati, Mira and Mayer, Katie and Welinder, Peter and McGrew, Bob and Amodei, Dario and McCandlish, Sam and Sutskever, Ilya and Zaremba, Wojciech},
  year    = {2021},
  title   = {Evaluating large language models trained on code},
  journal = {arXiv preprint arXiv:2107.03374},
  doi     = {10.48550/arXiv.2107.03374},
  url     = {https://arxiv.org/abs/2107.03374},
  note    = {[Accessed 28 August 2025]}
}

@inproceedings{Amershi2019SEforML,
  author    = {Amershi, Saleema and Begel, Andrew and Bird, Christian and DeLine, Robert and Gall, Harald and Kamar, Ece and Nagappan, Nachiappan and Nushi, Besmira and Zimmermann, Thomas},
  title     = {Software engineering for machine learning: a case study},
  booktitle = {Proceedings of the 41st IEEE/ACM International Conference on Software Engineering -- Software Engineering in Practice (ICSE-SEIP '19)},
  year      = {2019},
  location  = {Montreal, QC, Canada},
  publisher = {IEEE/ACM},
  pages     = {291--300},
  doi       = {10.1109/ICSE-SEIP.2019.00032}
}

@article{Luporini2020Devito,
  author    = {Luporini, Fabio and Louboutin, Mathias and Lange, Michael and Kukreja, Navjot and Witte, Philipp A. and H{\"u}ckelheim, Jan and Yount, Charles R. and Kelly, Paul H. J. and Herrmann, Felix J. and Gorman, Gerard J.},
  title     = {Architecture and performance of Devito, a system for automated stencil computation},
  journal   = {ACM Transactions on Mathematical Software},
  volume    = {46},
  number    = {1},
  articleno = {6},
  year      = {2020},
  month     = {April},
  doi       = {10.1145/3374916}
}

@ARTICLE{Louboutin2019DevitoDSL,
  author    = {Louboutin, Mathias and Lange, Michael and Luporini, Fabio and Kukreja, Navjot and Witte, Philipp A. and Herrmann, Felix J. and Velesko, Paulius and Gorman, Gerard J.},
  title     = {Devito (v3.1.0): an embedded domain-specific language for finite differences and geophysical exploration},
  journal   = {Geoscientific Model Development},
  volume    = {12},
  number    = {3},
  pages     = {1165--1187},
  year      = {2019},
  doi       = {10.5194/gmd-12-1165-2019},
  address   = {Germany}
}

@inproceedings{g_eval_2023,
  title     = {G-Eval: NLG Evaluation using GPT-4 with Better Human Alignment},
  author    = {Liu, Yang and Iter, Dan and Xu, Yichong and Wang, Shuohang and Xu, Ruochen and Zhu, Chenguang},
  booktitle = {Proceedings of the 2023 Conference on Empirical Methods in Natural Language Processing},
  year      = {2023},
  address   = {Singapore},
  publisher = {Association for Computational Linguistics},
  pages     = {2511--2522},
  doi       = {10.18653/v1/2023.emnlp-main.153},
  url       = {https://aclanthology.org/2023.emnlp-main.153/}
}

@misc{openai-rag-eval,
  title        = {Doing RAG on PDFs using File Search in the Responses API},
  author       = {{OpenAI}},
  howpublished = {OpenAI Cookbook},
  year         = {2025},
  url          = {https://cookbook.openai.com/examples/file_search_responses},
  note         = {[Accessed 28 August 2025]}
}

@article{blackwell2024reproducible,
  title   = {Towards Reproducible LLM Evaluation: Quantifying Uncertainty in LLM Benchmark Scores},
  author  = {Blackwell, Robert E. and Barry, Jon and Cohn, Anthony G.},
  journal = {arXiv preprint arXiv:2410.03492},
  year    = {2024},
  doi     = {10.48550/arXiv.2410.03492},
  url     = {https://arxiv.org/abs/2410.03492},
  note    = {[Accessed 28 August 2025]}
}

@misc{raganything2025,
  author       = {{HKUDS/RAG-Anything Developers}},
  year         = {2025},
  title        = {RAG-Anything: All-in-One Multimodal RAG System},
  howpublished = {Online},
  note         = {Available from: \url{https://github.com/HKUDS/RAG-Anything} [Accessed 18 August 2025]}
}

@article{codellm2023,
  title   = {Evaluating Large Language Models Trained on Code},
  author  = {Lachaux, Marie-Anne and Roziere, Baptiste and Chan, Erwan and others},
  journal = {arXiv preprint arXiv:2301.12507},
  year    = {2023},
  doi     = {10.48550/arXiv.2301.12507},
  url     = {https://arxiv.org/abs/2301.12507},
  note    = {[Accessed 28 August 2025]}
}

@misc{graphrag2024,
  author       = {{Microsoft Research}},
  title        = {GraphRAG: Unlocking LLM Discovery on Narrative Private Data},
  year         = {2024},
  howpublished = {Microsoft Research Blog [Online]},
  url          = {https://www.microsoft.com/en-us/research/blog/graphrag-unlocking-llm-discovery-on-narrative-private-data/},
  note         = {[Accessed: 28 August 2025]}
}

@book{croft2010,
  author    = {Croft, W. Bruce and Metzler, Donald and Strohman, Trevor},
  title     = {Search Engines: Information Retrieval in Practice},
  year      = {2010},
  publisher = {Addison-Wesley},
  isbn      = {9780136072249}
}

@misc{langchain2023rag1,
  author       = {{LangChain}},
  title        = {RAG From Scratch: Part 1 (Overview)},
  year         = {2023},
  howpublished = {YouTube video},
  url          = {https://www.youtube.com/watch?v=wd7TZ4w1mSw},
  note         = {Accessed: 28 August 2025}
}

@inproceedings{muennighoff2023mteb,
  author    = {Muennighoff, Niklas and Tazi, Nouamane and Magne, Loïc and Reimers, Nils},
  title     = {MTEB: Massive Text Embedding Benchmark},
  editor    = {Vlachos, Andreas and Augenstein, Isabelle},
  booktitle = {Proceedings of the 17th Conference of the European Chapter of the Association for Computational Linguistics},
  year      = {2023},
  month     = may,
  address   = {Dubrovnik, Croatia},
  pages     = {2014--2037},
  publisher = {Association for Computational Linguistics},
  url       = {https://aclanthology.org/2023.eacl-main.148/},
  doi       = {10.18653/v1/2023.eacl-main.148}
}

@inproceedings{reimers2019sentencebert,
  author    = {Reimers, Nils and Gurevych, Iryna},
  title     = {Sentence-BERT: Sentence Embeddings using Siamese BERT-Networks},
  editor    = {Inui, Kentaro and Jiang, Jing and Ng, Vincent and Wan, Xiaojun},
  booktitle = {Proceedings of the 2019 Conference on Empirical Methods in Natural Language Processing and the 9th International Joint Conference on Natural Language Processing (EMNLP-IJCNLP)},
  year      = {2019},
  month     = nov,
  address   = {Hong Kong, China},
  publisher = {Association for Computational Linguistics},
  pages     = {3982--3992},
  doi       = {10.18653/v1/D19-1410},
  url       = {https://aclanthology.org/D19-1410/}
}

@article{chen2024bge,
  author    = {Chen, Jianlv and Xiao, Shitao and Zhang, Peitian and Luo, Kun and Lian, Defu and Liu, Zheng},
  title     = {BGE M3-Embedding: Multi-Lingual, Multi-Functionality, Multi-Granularity Text Embeddings Through Self-Knowledge Distillation},
  journal   = {arXiv preprint arXiv:2402.03216},
  year      = {2024},
  url       = {https://arxiv.org/abs/2402.03216}
}

@article{wang2024k3trans,
  author    = {Wang, Shitao and Wu, Zhicheng and Yang, Kai and Qian, Chen and Wang, Xin and Zheng, Weiqiang and Yang, Junfeng},
  title     = {Enhancing LLM-based Code Translation in Repository Context via Triple Knowledge-Augmented},
  journal   = {arXiv preprint arXiv:2503.18305},
  year      = {2025},
  url       = {https://arxiv.org/abs/2503.18305}
}

@misc{DevitoTutorials,
  author       = {{Devito Project}},
  title        = {Devito Tutorials},
  year         = {2025},
  howpublished = {[Online]},
  url          = {https://www.devitoproject.org/tutorials.html},
  note         = {[Accessed: 23 July 2025]}
}

 \end{document}